\documentclass[11pt]{article}
\usepackage[utf8]{inputenc}
\usepackage[T1]{fontenc}
\usepackage{graphicx}
\usepackage{longtable}
\usepackage{wrapfig}
\usepackage{rotating}
\usepackage[normalem]{ulem}
\usepackage{amsmath}
\usepackage{amssymb}
\usepackage{capt-of}
\usepackage{hyperref}
\usepackage[T1]{fontenc}
\usepackage{lmodern}
\usepackage{csquotes}
\usepackage[font=small,labelfont=bf, justification=justified, format=plain]{caption}
\author{J.-M. Chauvet}
\date{\today}
\title{Memory Traces: \\Are Transformers Tulving Machines?}
\hypersetup{
 pdfauthor={J.-M. Chauvet},
 pdftitle={Memory Traces: Are Transformers Tulving Machines?},
 pdfkeywords={},
 pdfsubject={},
 pdfcreator={Emacs 29.3 (Org mode 9.6.15)}, 
 pdflang={English}}
\makeatletter
\newcommand{\citeprocitem}[2]{\hyper@linkstart{cite}{citeproc_bib_item_#1}#2\hyper@linkend}
\makeatother

\usepackage[notquote]{hanging}
\begin{document}

\maketitle

\section*{Abstract}
\label{sec:org9a00d17}
Memory traces--changes in the memory system that result from the perception and encoding of an event--were measured in pioneering studies by Endel Tulving and Michael J. Watkins in 1975 \citeprocitem{1}{[1]}. These and further experiments informed the maturation of Tulving's memory model, from the GAPS (\emph{General Abstract Processing System}) to the SPI (\emph{Serial-Parallel Independent}) model \citeprocitem{2}{[2]}, \citeprocitem{3}{[3]}. Having current top of the line LLMs revisit the original Tulving-Watkins tests may help in assessing whether foundation models completely instantiate or not this class of psychological models.

\section*{Introduction}
\label{sec:org436e638}
In the course of almost 50 years, Endel Tulving (1927-2023) developed a model of human memory, based on carefully designed experiments, culminating with the Serial-Parallel Independent model (SPI) \citeprocitem{3}{[3]}. In this popular non-computational model, memory is composed of several hierarchically organized systems and the functional relationships between these systems drive its performance. The \emph{encoding} is serial: successful coding and storage in a system depends on successful coding and storage in lower systems in the hierarchy. The \emph{retrieval} from a system, in contrast, is independent of retrievals from the other systems storing the information in parallel. We designate this model and its variations and derivatives \citeprocitem{4}{[4]} by the term \emph{Tulving Machine} models.

\begin{figure}[htbp]
\centering
\includegraphics[width=.9\linewidth]{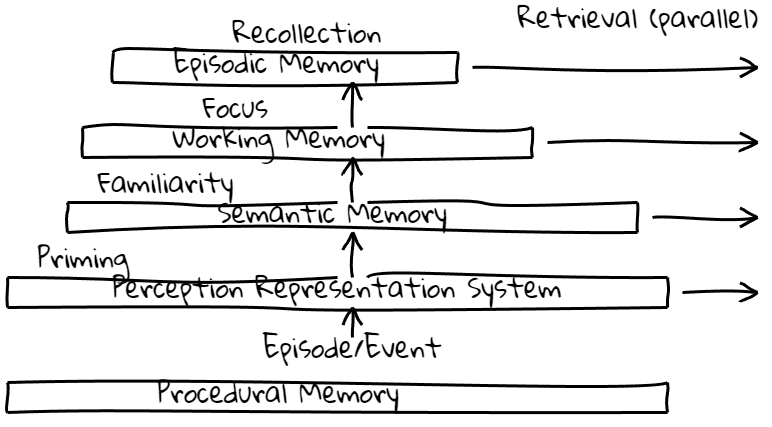}
\caption{\label{fig:org70ac46a}\textbf{The SPI Model of memory.} The original 3-layer GAPS model later expanded to 5 systems \citeprocitem{5}{[5]} into the Serial-Parallel Independent model. The \emph{Procedural Memory} sits apart from a hierarchy of four systems: a \emph{Perception Representation System} linked to \emph{priming} effects, \emph{Semantic Memory} associated with familiarity, \emph{Working Memory} acting as a buffer \citeprocitem{6}{[6]} to the \emph{Episodic Memory} associated with \emph{recollection}, ``the individual's subjective awareness of remembering'' \citeprocitem{2}{[2]}. Encoding of an event or episode happens serially along the vertical arrows, and information (memory traces) are stored in parallel in each system. Retrievals from each system, on the other hand, happens independently from other systems (horizontal arrows). Source for image: \citeprocitem{5}{[5]}.}
\end{figure}

In this study, we are interested in the properties of memory traces in LLMs, compared to similar findings in human memory performance tests. More specifically, we adapt the test protocol of Tulving and Watkins \citeprocitem{1}{[1]} to operate on LLMs. Interestingly enough, the original
idea informing the test, e.g. thinking of memory trace as a ``cognitive blueprint'' that specifies the conditions under which the recollection of an event will occur, rather than only as an after-effect of the registration of the event, seems particurlary relevant to prompt engineering in LLM operations. In both GAPS and SPI models, (a) the memory of an input event or episode is not a copy of the event, (b) the notion of a fixed trace is reconciled with the observation that its activation will depend on the retrieval environment (cue) and (c) the relation between input conditions and the trace of an item are actually observable. Namely the memory trace is defined operationally in terms of the relations between the queries directed at the system, here the LLM, and the output from the system, given a particular encoding of the event \citeprocitem{2}{[2]}. Using prompts both for the encoding, i.e. the registration of an event, and later retrievals, we make sure that effective retrieval cues might be quite different from the original event in a disciplined way. The test measures comparative effectiveness of various retrieval cues in producing recall.

Retrieval cueing is a prime tool for gaining information about the composition of memory traces. In the Tulving-Watkins experiment, a given trace is successively probed with two (or more) different types of retrieval cues. In the original paper, a \emph{reduction method} is introduced to yield an observation of the relations among retrieval cues with respect to the trace. Central to the reduction method is Tulving's principle of \emph{encoding specificity} \citeprocitem{2}{[2]}: a retrieval cue is effective to the extent that its informational contents match the informational content of the trace. The quantitative measure of the effectiveness of a retrieval cue enables us to infer informational properties of a particular trace and to estimate quantitatively its constituent elements. Tulving suggests the name \emph{valence}, with respect to a particular trace, for the effectiveness of a retrieval cue, i.e. the probability with which that event can be recalled in the presence of the cue. This is the \emph{gross valence}; in foresight of information theory \citeprocitem{7}{[7]}, Tulving also introduced the \emph{reduced valence} of cue \emph{X} by \emph{Y} as the probability that the target can be recalled to cue \emph{X} and not to cue \emph{Y}, and the \emph{common valence} of two cues, \emph{X} and \emph{Y}, as the probability that the target can be retrieved by cue \emph{X} as well as by cue \emph{Y}. (In modern parlance the term is \emph{mutual information}.) The Tulving-Watkins test is built around the simple operations of presenting a word, the event or the episode to be remembered in this test, and probing it with two successive cues, repeated many times in order to estimate the valences of the cues.

Simple information-theoretic considerations, termed reduction method by Tulving and Watkins, yield estimations for valences of retrieval cues from the data points from the experimental test. We compare the observations on human subjects from the original papers to similar observations on LLMs (see Section \hyperref[sec:org7587c07]{Methods}) and discuss some implications for the understanding of remembering in LLMs.

\section*{Results}
\label{sec:org754e25b}
\begin{figure}[htbp]
\centering
\includegraphics[width=.9\linewidth]{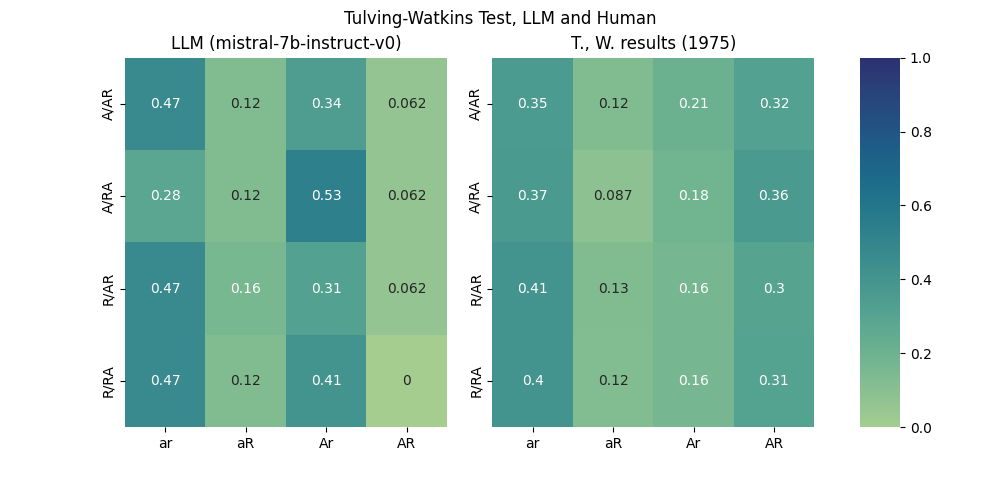}
\caption{\label{fig:orgffc6acc}\textbf{Summary of memory performance (recollection) in LLM, left, and human, right.} Both tables are presented in the original \emph{data matrix} format from \citeprocitem{1}{[1]}, showing frequency of occurrences normalized to \emph{[0,1]}. Following the original paper, the test targets two independent categories of word-cues: \emph{associative}, noted \emph{A}, and \emph{rhyming}, noted \emph{R} (see text). In a test session a list of words, each given with a context word, either \emph{A} or \emph{R}, is memorized. Two cued-retrievals are then performed in sequence, in the same LLM chat, exploring both alternations of cue categories, \emph{A} then \emph{R} and \emph{R} then \emph{A}. This setup of two encodings and two retrievals results in four rows marked on the vertical axis as A/AR, A/RA, R/AR, R/RA. In a table, each row displays frequency of occurrences of the four possible outcomes of the Tulving-Watkins cued dual retrieval tests: the word has been retrieved (upper case letter \emph{A} or \emph{R}) from the cue, or not (lower-case letter \emph{a} or \emph{r}). Drastic differences between LLM and human in memory performance profiles are discussed in the text.}
\end{figure}

An overview of memory performance of the \texttt{mistral-7b-instruct-v0} LLM is presented in Figure  \ref{fig:orgffc6acc} (left) for direct comparison with results discussed in \citeprocitem{1}{[1]} (right). A first remark is that overall the probability of failure to pass both recall tests is higher in LLMs than in human subjects, except for the A/RA signature--associated word context in encoding, followed by rhyming then associative retrieval cues in recall. The rhyming cue seems more distracting for the human subject than for the LLM, or rather as distracting as the other way around in human subjects' responses.

The second remark is that LLMs are much better than human subjects at remembering through associative retrieval cues rather than through rhyming cues (comparing third column from the left in each table). Human subjects display a more balanced performance remembering both through associative and rhyming retrieval cues (comparing the third and fourth columns in each table).

Finally, human subjects perform best on associative encoding, recalling the target word on both orders of retrieval cues with probabilities \emph{.32} and \emph{.36}, about one in three. LLMs, in contrast, while similarly slightly better on associative encoding are much better at associative recall, with probability \emph{.53}, passing over half of the tests.

\begin{table}[htbp]
\caption{\label{tab:org986db74}\textbf{Valence Trace Matrices in LLMs, left, and in human, right. Associative encoding, top and rhyming encoding, bottom.} Gross valences for each encoding-retrieval cue pair is displayed in italics, under or next to the `Total' heading.}
\centering
\begin{tabular}{cc|ccc|c|ccc|}
Retrieval &  & R &  &  &    & R &  & \\[0pt]
Cue &  & + & - & Total &    & + & - & Total\\[0pt]
\hline
\hline
A & + & 0.01 & 0.47 & \emph{0.49} &  & 0.33 & 0.19 & \emph{0.52}\\[0pt]
 & - & 0.10 & 0.42 &  &  & 0.12 & 0.36 & \\[0pt]
 & Total & \emph{0.11} &  &  &  & \emph{0,45} &  & \\[0pt]
\hline
 &  &  &  &  &  &  &  & \\[0pt]
\hline
Cue &  & + & - & Total &    & + & - & Total\\[0pt]
\hline
\hline
A & + & 0.02 & 0.36 & \emph{0.38} &  & 0.27 & 0.16 & \emph{0.43}\\[0pt]
 & - & 0.12 & 0.50 &  &  & 0.16 & 0.41 & \\[0pt]
 & Total & \emph{0.14} &  &  &  & \emph{0,43} &  & \\[0pt]
\hline
\end{tabular}
\end{table}

A preliminary remark is appropriate as we look into Table \ref{tab:org986db74}. The reduction method used by Tulving and Watkins assumes that in successive retrievals a cue that does not elicit a recall leaves the memory trace unchanged. In the \hyperref[sec:org7587c07]{Methods} section this is captured by equation \ref{eq:orgba63194} which must hold. While this is the case, up to some negligible noise or sampling variability in the human performance reported in the original paper, this is no longer the case in the measured LLM memory performance. We apply nonetheless the original reduction method, even though the measured discrepancy in the results shown in Table \ref{tab:org986db74} is \emph{.08} and \emph{.17}, respectively for associative and rhyming encodings, both values not negligible in our opinion (\emph{.06}, however, for aggregate encodings, which is equal to the discrepancy found in the results published in \citeprocitem{1}{[1]}).

The apparent distinction between human and LLM performances is that LLM recall, although slightly inferior in associative cue retrieval, falls comparatively far behind in rhyming cue retrievals. Failure to recall any of the target words is higher in LLMs, by \emph{.08} and \emph{.09} for respective encodings, and successful recall on both encodings is almost null for LLMs while, in contrast, it is the highest probable result in humans when recall occurs.

\section*{Discussion}
\label{sec:orgcbec999}
Although LLM neural networks, and particularly their \emph{Transformer} architecture, were not specifically designed for memory tasks as other networks may have been \citeprocitem{8}{[8]}, \citeprocitem{9}{[9]}, the results above and those in \citeprocitem{10}{[10]} show their episodic memory performance not to be insignificant. As ``simple-minded''-- \emph{dixit} Tulving -- at it may sound, the idea of describing memory traces with information-theoretic notions of gross, common and reduced valences of two or more retrieval cues was proved successful in exploring human memory performance \citeprocitem{5}{[5]}. This principle also proves its value at outlining what a memory trace looks like in a chat with a Transformer-based LLM -- if only because (i) it provides an internally consistent set of statements about what the memory trace is and what its properties are, and (ii) a set of objectively specified rules for translating experimental observations into theoretical terms in both human and LLM subjects. Needless to say, more investigations along other dimensions of retrieval cues, beyond generic association and rhyming, would be required to specify the said properties.

As Tulving noted, this outline neither demands nor assumes any kind of \emph{real} existence of some \emph{thing} stored in the memory system (whether human or LLM for that matter). The valence-theoretical memory trace is a hypothetical symbolic construct that serves to organize a large list of relations between different questions directed at the system and the output of the same system. Recently it was shown that, similarly, a large subset of the relations between words, which constitute much of the knowledge encoded in pretrained LLMs, are well-approximated by a single linear transformation \citeprocitem{11}{[11]}. In the Tulving Machine Model, this voluminous relational knowledge plays the role of the Semantic Memory, which is linked to familiarity responses.

Peculiarities, of course, emerge from the comparison of valences observed in human and in LLM subjects (see \hyperref[sec:org754e25b]{Results}). Further psychological studies \citeprocitem{12}{[12]}, \citeprocitem{13}{[13]} investigated the relation between episodic and semantic memory, providing a framework in which the results above hint at a different nature of memory in humans and in LLMs. This line of psychometric research work convincingly demonstrated that ``patients with semantic dementia, who have an incontrovertible deficit in semantic memory, are reported to show good day-to-day memory for recent events; but experimental evidence on their anterograde episodic memory/new learning is somewhat sparse and does not always tell a consistent story.'' Obviously, the situation is somewhat inverted in the case of LLMs: namely, many of the recalls fail not because there is no target word recollected but because the LLM response involves an extra-list word. Words strongly associated to the retrieval cue in semantic memory are produced in response, whatever the cue's valence is relative to the target. Relations in the pretrained semantic memory of LLMs often seem to overwhelm and supersede locally memorized relations in a chat.

Another famous case, patient K.C. \citeprocitem{14}{[14]}, is evidence that ``[K.C's] ability to make use of knowledge and experiences from the time before his accident shows a sharp dissociation between semantic and episodic memory. A good deal of his general knowledge of the world, including knowledge about himself, is preserved, but he is incapable of recollecting any personally experienced events. In displaying such episodic amnesia, which encompasses an entire lifetime of personal experiences, K.C. differs from many other amnesic cases.'' While superficially similar in the poor LLM results at recollection (and specifically so in non associative encodings circumstances) compared to human memory tasks performance, patient K.C.'s predicament is much stronger.

Where then do Transformers sit, between semantic dementia and episodic amnesia?

Elaborations of the interactions between episodic and semantic memory, from neuroimaging, led to adding to the Tulving Machine an \emph{episodic buffer} \citeprocitem{6}{[6]}, posited as ``a limited capacity system that provides temporary storage of information held in a multi-modal code, which is capable of binding information from the subsidiary systems [working and episodic memory], and from long-term memory, into a unitary episodic representation. Conscious awareness is assumed to be the principal mode of retrieval from the buffer.'' The relevance of Baddeley's proposal to LLM understanding may be read in the following statement: ``The suggestion that the episodic buffer forms the crucial interface between memory and conscious awareness places it at the centre of the highly active line of research into the role of phenomenological factors in memory and cognition. Tulving, for example, defines his concept of episodic memory explicitly in terms of its associated phenomenological experience of remembering. Although not all theorists would wish to place phenomenological experience so centrally, there is increasing evidence to suggest that conscious monitoring of the evidence supporting an apparent memory plays a crucial role in separating accurate recall from false memory, confabulation and delusion.'' The various roles played by conscious memory \citeprocitem{15}{[15]} and its strong candidate neural substrates, chiefly the hippocampus and the medial temporal lobes \citeprocitem{16}{[16]}, cite: \citeprocitem{17}{[17]}, are evidently without equivalent in LLM's \emph{artificial} neural networks, but would that ruin any effort to consolidate the Tulving Machine into being a guiding light in understanding LLMs' memory performance?

On the contrary, the Tulving Machine, now generally considered a valid human memory model, both on behavioral experimental psychology and on neurological grounds \citeprocitem{4}{[4]}, doubles as providing quantitative instruments to investigate the distinctive relationships between its different memory \emph{components} (episodic, working, semantic) when looking at LLMs. Within this model, the above results, in light of Baddeley's episodic buffer function, maybe interpreted as LLM having no, or little, ability to identify the source of remembered information, which is critical for many cognitive tasks \citeprocitem{18}{[18]}. Indeed, a central claim of the so-called \emph{source monitoring} approach is that people do not typically \emph{directly} retrieve a label that specifies a memory's source--a variety of characteristics that, collectively, specify the conditions under which a memory is acquired, i.e. its specific encoding at acquisition time--rather, as Tulving delineated \citeprocitem{2}{[2]}, activated memory records are evaluated and attributed to particular sources through \emph{decision processes} performed during remembering \citeprocitem{19}{[19]}, \citeprocitem{20}{[20]}, \citeprocitem{21}{[21]}. Tools like cue valences can help design experimental protocols intended to focus attention on the LLM processes involved in distinguishing between the recalled and the self-generated in memory. The overall picture that emerges from the initial evidence reviewed here and in \citeprocitem{10}{[10]} is one of a LLM memory system which, in stark contrast to human memory, does not preserve information about the origin of information remarkably well. We feel that the Tulving Machine framework is viable and  warrants further investigation, and specifically on how this information about the origin of information is filtered in the decision process attributing memories to sources.

\section*{Methods}
\label{sec:org7587c07}
We transpose the successive probes and reduction methods described in \citeprocitem{1}{[1]} for testing LLM ``subjects''. Individual experiments are programmed as Python scripts interacting with LLMs through the LLM CLI utility and library \citeprocitem{22}{[22]} (Python 3.11.8 on Windows 10). Results presented and discussed in this paper were obtained with \texttt{mistral-7b-instruct-v0} \citeprocitem{23}{[23]}. (Results with smaller models, e.g. \texttt{orca-mini-3b} \citeprocitem{24}{[24]}, were not significant or reliable enough.)

Sixteen unrelated to-be-remembered words were selected from the Oxford English Dictionary. Each of them was presented to a LLM in a prompt asking for an associated word and a rhyming word. While the LLM responses were correct for associated words, it was not so for rhyming words. Rhyme cue words were then obtained using the CMU Pronouncing Directory \citeprocitem{25}{[25]}. This selection was repeated to produce a list of sixteen target words, each with two associated words and two rhyming words to be used as cues in the test.

In a chat, the LLM was presented with the target words, each accompanied with a context word, the \emph{input cue}. One half of the context words were associatively (\emph{A}) related to the target word, while the other half were words that rhymed (\emph{R}) with the target word. The two types of input cues, \emph{A} and \emph{R}, defined the two encoding conditions of the experiment. In this presentation, the context was stated explicitly either as ``is associated to'' or as ``rhymes with''. The firts prompt in the chat instructed the LLM to memorize the list and prepare for a cue-recall test.

In these tests, implemented as question prompts in the same chat, each target word was probed twice successively, once with an associative cue--different from the one used in the memorization of the list-- and once with a rhyming cue--also different from the one in the given context of the initial list. To accommodate various token-sized contexts, the 16 words were tested in four batches of four chats, rather than in one longer 16-word chat.

Of each of the subset of 4 targets, two had been encoded with respect to an associative context word and two with respect to a rhyming word. One word in each of these pair of targets was first probed with a rhyming cue and then with an associative cue (\emph{RA}), while for the other word, the order of cues was reversed (\emph{AR}). Therefore each target word within the list represented a unique set of experimental conditions.

The results of these observations, i.e. the LLM producing or not the target word, were collected into data matrices which were then treated by the reduction method.

In the Tulving Model, the pattern of cue valences (say for cue \emph{X} and \emph{Y}) shown in Table \ref{tab:org4582fb1} describes the trace of the target word.

\begin{table}[htbp]
\caption{\label{tab:org4582fb1}\textbf{Data Matrix: Gross, Common and Reduced valences of retrieval cue \emph{X} followed by \emph{Y}}. Notations: + means the target word is recognized when retrieval cue is used, - if not. The number of occurrences of possible results in each table cell is normalized to its frequency in \emph{[0,1]} as an estimate of its probability. Results are displayed as the two ordered retrieval cues, capital letters when the target word is recognized when the cue is presented, lower-case letters when the word is not recognized when the cue is presented.}
\centering
\begin{tabular}{cccc}
  & Second cue \emph{Y}  &   &  \\[0pt]
First cue \emph{X}  & + & - & Total \\[0pt]
\hline
+ & XY & Xy  & X\\[0pt]
- & xY & xy  & x\\[0pt]
Total & Y & y & \\[0pt]
\hline
\end{tabular}
\end{table}

The table collects the notional valences defined by Tulving and Watkins. The gross valence of cue \emph{X} is simply \emph{X = XY + Xy}, where the convention is to use lower-case letter for negative recall and upper-case letter for positive recall of the target word. The gross valence of cue \emph{Y} is similarly read as \emph{Y = XY + xY}. The reduced valences are shown in each individual cell, e.g. the valence of cue \emph{X} reduced by cue \emph{Y} is \emph{Xy}. Finally the common valence of cue \emph{X} and cue \emph{Y} is in the upper-left cell: \emph{XY}.

The quantitative description of memory traces is however a bit more complicated, as the valences of the two cues may depend on the order in which they are actually presented. The order of presentation \emph{YX}, for instance, would yield another data matrix shaped as Table \ref{tab:org4582fb1}. And this is to be expected if the first cue changed, or re-encoded, the trace, since the two cues would practically be applied to different re-encoded traces. Tulving and Watkins developed the \emph{reduction method} to compute (in an information theory compliant way) a proper \emph{trace matrix} from the two \emph{XY} and \emph{YX} data matrices.

The reduction method, for the details of which we refer to the original paper \citeprocitem{1}{[1]}, relies on the basic assumption that presenting a retrieval cue might cause a re-encoding of of the target trace only if the cue is successful in effecting a recall; a cue that does not elicit a recall of the target item is assumed to leave the trace intact. This assumption, in relation to LLMs, is further discussed in the previous sections. Note that the assumption entails that \emph{xy} and \emph{yx} are equal, barring noise and measurement errors, since none of the cues, whichever the order of presentation, elicited a recall of the target word. Provided the assumption, cells in the trace matrix are defined by the following elementary calculations:

\begin{equation}
\label{eq:org3f95017}
x_t y_t = (xy + yx)/2
\end{equation}

\begin{equation}
\label{eq:org934c231}
X_t = X(1 - x_t y_t)/(1 - xy)
\end{equation}

\begin{equation}
\label{eq:org78e1bf8}
Y_t = Y(1 - x_t y_t)/(1 - yx)
\end{equation}

\begin{equation}
\label{eq:org85940c6}
x_t Y_t = xY(1 - x_t y_t)/(1 - xy)
\end{equation}

\begin{equation}
\label{eq:orgbb351cd}
X_t y_t = yX(1 - x_t y_t)/(1 - yx)
\end{equation}

under the assumption above, captured by the following equation:

\begin{equation}
\label{eq:orgba63194}
X_t Y_t + X_t y_t + x_t Y_t = Y_t X_t + Y_t x_t + y_t X_t
\end{equation}

with the same upper-case/lower-case notation for the trace matrix gross valences, \(X_t\) and \(Y_t\), reduced valences, \(X_t y_t\) and \(x_t Y_t\), and common valences, \(X_t Y_t\), irrespective of the order.

\begin{table}[htbp]
\caption{\label{tab:org28b3c81}The trace matrix obtained by applying Tulving and Watkins reduction method to both data matrices from retrieval sequences \emph{XY} and \emph{YX}. The valences, irrespective of the order of presentation, are calculated by equations \ref{eq:org3f95017} to \ref{eq:orgbb351cd}, as indicated. Residual values are obtained by row or column additions.}
\centering
\begin{tabular}{llll}
  &  Cue \emph{Y}  &   &  \\[0pt]
 Cue \emph{X}  &  +  &  -  &  Total \\[0pt]
\hline
 + &  \(X_t Y_t\) (subtract from total) &  \(X_t y_t\) (from Eqn. \ref{eq:orgbb351cd})  &  \(X_t\) (from Eqn. \ref{eq:org934c231})\\[0pt]
 - &  \(x_t Y_t\)   (from Eqn. \ref{eq:org85940c6}) &  \(x_t y_t\) (from Eqn. \ref{eq:org3f95017})  &  \(x_t\) (add row)\\[0pt]
 Total &  \(Y_t\) (from Eqn. \ref{eq:org78e1bf8}) &  \(y_t\) (add column) &  \\[0pt]
\hline
\end{tabular}
\end{table}

\section*{References}
\label{sec:orgd8d5fbb}

\noindent \hypertarget{citeproc_bib_item_1}{[1] E. Tulving and M. J. Watkins, “Structure Of Memory Traces,” \textit{Psychological review}, vol. 82, no. 4, pp. 261–275, Jul. 1975, doi: \href{https://doi.org/10.1037/h0076782}{10.1037/h0076782}.}

\noindent \hypertarget{citeproc_bib_item_2}{[2] E. Tulving, \textit{Elements of Episodic Memory}. Oxford University Press, 1983.}

\noindent \hypertarget{citeproc_bib_item_3}{[3] E. Tulving, “Organization of memory: Quo vadis,” in \textit{The cognitive neurosciences}, M. S. Gazzaniga, Ed., MIT Press, 1995, pp. 839–847.}

\noindent \hypertarget{citeproc_bib_item_4}{[4] F. Eustache and B. Desgranges, “MNESIS: towards the integration of current multisystem models of memory,” \textit{Neuropsychol. rev.}, vol. 18, no. 1, pp. 53–69, Mar. 2008.}

\noindent \hypertarget{citeproc_bib_item_5}{[5] B. Desgranges and F. Eustache, “Les conceptions de la mémoire déclarative d’endel tulving et leurs conséquences actuelles,” \textit{Revue de neuropsychologie}, vol. Volume 3, no. 2, pp. 94–103, Jun. 2011, doi: \href{https://doi.org/10.1684/nrp.2011.0169}{10.1684/nrp.2011.0169}.}

\noindent \hypertarget{citeproc_bib_item_6}{[6] A. Baddeley, “The episodic buffer: a new component of working memory?,” \textit{Trends in cognitive sciences}, vol. 4, no. 11, pp. 417–423, Nov. 2000, doi: \href{https://doi.org/10.1016/s1364-6613(00)01538-2}{10.1016/s1364-6613(00)01538-2}.}

\noindent \hypertarget{citeproc_bib_item_7}{[7] T. M. Cover and J. A. Thomas, \textit{Elements of information theory}. Wiley, 2001. doi: \href{https://doi.org/10.1002/0471200611}{10.1002/0471200611}.}

\noindent \hypertarget{citeproc_bib_item_8}{[8] S. Hochreiter and J. Schmidhueber, “Long short-term memory,” \textit{Neural comput.}, vol. 9, no. 8, pp. 1735–1780, 1997, doi: \href{https://doi.org/10.1162/NECO.1997.9.8.1735}{10.1162/NECO.1997.9.8.1735}.}

\noindent \hypertarget{citeproc_bib_item_9}{[9] J. Weston, S. Chopra, and A. Bordes, “Memory networks.” 2014.}

\noindent \hypertarget{citeproc_bib_item_10}{[10] J.-M. Chauvet, “Memory GAPS: Would LLMs pass the Tulving Test?” [Online]. Available: \url{https://arxiv.org/abs/2402.16505}}

\noindent \hypertarget{citeproc_bib_item_11}{[11] E. Hernandez \textit{et al.}, “Linearity of Relation Decoding in Transformer Language Models.”}

\noindent \hypertarget{citeproc_bib_item_12}{[12] A.-L. Adlam, K. Patterson, and J. Hodges, “‘i remember it as if it were yesterday’: Memory for recent events in patients with semantic dementia,” \textit{Neuropsychologia}, vol. 47, no. 5, pp. 1344–1351, Apr. 2009, doi: \href{https://doi.org/10.1016/j.neuropsychologia.2009.01.029}{10.1016/j.neuropsychologia.2009.01.029}.}

\noindent \hypertarget{citeproc_bib_item_13}{[13] K. S. Graham, J. S. Simons, K. H. Pratt, K. Patterson, and J. R. Hodges, “Insights from semantic dementia on the relationship between episodic and semantic memory,” \textit{Neuropsychologia}, vol. 38, no. 3, pp. 313–324, Mar. 2000, doi: \href{https://doi.org/10.1016/s0028-3932(99)00073-1}{10.1016/s0028-3932(99)00073-1}.}

\noindent \hypertarget{citeproc_bib_item_14}{[14] R. S. Rosenbaum \textit{et al.}, “The case of k.c.: contributions of a memory-impaired person to memory theory,” \textit{Neuropsychologia}, vol. 43, no. 7, pp. 989–1021, 2005, doi: \url{https://doi.org/https://doi.org/10.1016/j.neuropsychologia.2004.10.007}.}

\noindent \hypertarget{citeproc_bib_item_15}{[15] M. A. Conway, A. F. Collins, S. E. Gathercole, and S. J. Anderson, “Recollections of true and false autobiographical memories.,” \textit{Journal of experimental psychology: General}, vol. 125, no. 1, pp. 69–95, Mar. 1996, doi: \href{https://doi.org/10.1037/0096-3445.125.1.69}{10.1037/0096-3445.125.1.69}.}

\noindent \hypertarget{citeproc_bib_item_16}{[16] R. Jaffard, “La mémoire déclarative et le modèle de squire,” \textit{Revue de neuropsychologie}, vol. Volume 3, no. 2, pp. 83–93, Jun. 2011, doi: \href{https://doi.org/10.1684/nrp.2011.0174}{10.1684/nrp.2011.0174}.}

\noindent \hypertarget{citeproc_bib_item_17}{[17] A. Yonelinas, C. Hawkins, A. Abovian, and M. Aly, “The role of recollection, familiarity, and the hippocampus in episodic and working memory,” May 2023, doi: \href{https://doi.org/10.31234/osf.io/5gwz9}{10.31234/osf.io/5gwz9}.}

\noindent \hypertarget{citeproc_bib_item_18}{[18] K. J. Mitchell and M. K. Johnson, “Source monitoring: Attributing mental experiences,” in \textit{The oxford handbook of memory}, Oxford University PressNew York, NY, 2000, pp. 179–196. doi: \href{https://doi.org/10.1093/oso/9780195122657.003.0012}{10.1093/oso/9780195122657.003.0012}.}

\noindent \hypertarget{citeproc_bib_item_19}{[19] S. A. Mutter and S. Hashtroudi, \textit{Cognitive effort requirements in recall, recognition, and lexical decision}. Defense Technical Information Center, 1985. doi: \href{https://doi.org/10.21236/ada170386}{10.21236/ada170386}.}

\noindent \hypertarget{citeproc_bib_item_20}{[20] S. Hashtroudi, S. A. Ferguson, V. A. Rappold, and L. D. Chrosniak, “Data-driven and conceptually driven processes in partial-word identification and recognition.,” \textit{Journal of experimental psychology: Learning, memory, and cognition}, vol. 14, no. 4, pp. 749–757, 1988, doi: \href{https://doi.org/10.1037/0278-7393.14.4.749}{10.1037/0278-7393.14.4.749}.}

\noindent \hypertarget{citeproc_bib_item_21}{[21] M. K. Johnson and C. L. Raye, “Reality monitoring.,” \textit{Psychological review}, vol. 88, no. 1, pp. 67–85, Jan. 1981, doi: \href{https://doi.org/10.1037/0033-295x.88.1.67}{10.1037/0033-295x.88.1.67}.}

\noindent \hypertarget{citeproc_bib_item_22}{[22] S. Willison, “LLM.” https://llm.datasette.io/en/stable/index.html, 2023.}

\noindent \hypertarget{citeproc_bib_item_23}{[23] A. Q. Jiang \textit{et al.}, “Mistral 7B.”}

\noindent \hypertarget{citeproc_bib_item_24}{[24] P. Mathur, “An explain tuned OpenLLaMA-3b model on custom wizardlm, alpaca, and dolly datasets,” \textit{Github repository, huggingface repository}. https://github.com/pankajarm/wizardlm\_alpaca\_dolly\_orca\_open\_llama\_3b, https://https://huggingface.co/psmathur/wizardlm\_alpaca\_dolly\_orca\_open\_llama\_3b; GitHub, HuggingFace, 2023.}

\noindent \hypertarget{citeproc_bib_item_25}{[25] Carnegie Mellon Speech Group, “The CMU Pronouncing Dictionary.” }\bigskip

\section*{Author information}
\label{sec:orgcd49ebb}
Jean-Marie Chauvet is a co-founder of Neuron Data and served as its CTO (1985-2000). He no longer maintains any affiliation.

J.-M. C. performed all analyses and wrote the manuscript as an independent researcher.

\section*{Ethics declarations}
\label{sec:org769f335}
The author declare no competing interests.

\section*{Electronic supplementary material}
\label{sec:orge5516b1}
\subsection*{Data Availability}
\label{sec:orgfd5e1d9}
Results of the Tulving-Watkins Tests analysed in the paper are publicly available in the repository:
\url{https://github.com/CRTandKDU/TulvingTest/tree/main/tulving/output}

\subsection*{Code Availability}
\label{sec:org6385109}
Python scripts for the Tulving Test and tabulation of their results are publicly available in the repository:
\url{https://github.com/CRTandKDU/TulvingTest/tree/main/tulving}
\end{document}